\documentclass[a4paper]{cas-dc}

\usepackage[numbers]{natbib}
\usepackage{subfigure}
\usepackage[justification=centering]{caption}
\usepackage{threeparttable}
\usepackage{caption}
\newcommand{\tabincell}[2]{\begin{tabular}{@{}#1@{}}#2\end{tabular}}
\usepackage{caption}
\usepackage{amsmath}
\usepackage{amssymb}
\usepackage{bm}
\usepackage{stfloats}
\usepackage{url}
\def\tsc#1{\csdef{#1}{\textsc{\lowercase{#1}}\xspace}}
\tsc{WGM}
\tsc{QE}
\tsc{EP}
\tsc{PMS}
\tsc{BEC}
\tsc{DE}

\begin{document}
	
\title [mode = title]{Robust Line Segments Matching via Graph Convolution Networks}                      

\let\WriteBookmarks\relax
\def\floatpagepagefraction{1}
\def\textpagefraction{.001}
\shorttitle{Robust Line Segments Matching via Graph Convolution Networks}
\shortauthors{QuanMeng Ma et~al.}
\author{QuanMeng Ma}
\address{School of Telecommunication Engineering, Xidian University, Xi'an, China.}

\author{Guang Jiang}[type=editor,
orcid=0000-0001-9363-2012]
\cormark[1]
\author{DianZhi Lai}

\cortext[cor1]{Corresponding author}

\nonumnote{qmma@stu.xidian.edu.cn(Q.Ma),gjiang@mail.xidian.edu.cn(G.Jiang)}

\begin{abstract}
Line matching plays an essential role in structure from motion (SFM) and simultaneous localization and mapping (SLAM), especially in low-textured and repetitive scenes. In this paper, we present a new method of using a graph convolution network to match line segments in a pair of images, and we design a graph-based strategy of matching line segments with relaxing to an optimal transport problem. In contrast to hand-crafted line matching algorithms, our approach learns local line segment descriptor and the matching simultaneously through end-to-end training. The results show our method outperforms the state-of-the-art techniques, and especially, the recall is improved from 45.28\% to 70.47\% under a similar presicion. The code of our work is available at \url{https://github.com/mameng1/GraphLineMatching}.
\end{abstract}



\maketitle

\section{Introduction}
Finding corresponding line or point features in images generally is the first step in many computer vision applications, such as structure from motion (SFM) and simultaneous localization and mapping (SLAM). Compared with line matching, point matching has been well studied, and many learning-based and hand-crafted methods have been proposed in the past two decades, so it is widely used in the fields of SLAM and SFM. In indoor scenes that lack texture, there may be very few feature points that can be detected. Due to the lack of distinguishing characteristics, it is challenging to match key points in these scenes. Unlike the points, line segments are usually located at the intersection of planes and edges of an object, so they can be detected more easily even in low-textured interiors. Besides, the lines can easily reflect the outline of the indoor scenes, so it is simpler to reconstruct the 3d structure of indoor scenes using line segments instead of points. So some researchers \cite{salaun2017line, li2017line, pumarola2017pl, hofer2017efficient} used line segments to reconstruct interiors and improve the robustness of SLAM system in the past several years.

\begin{figure}
	\centering
	\includegraphics[scale=0.45]{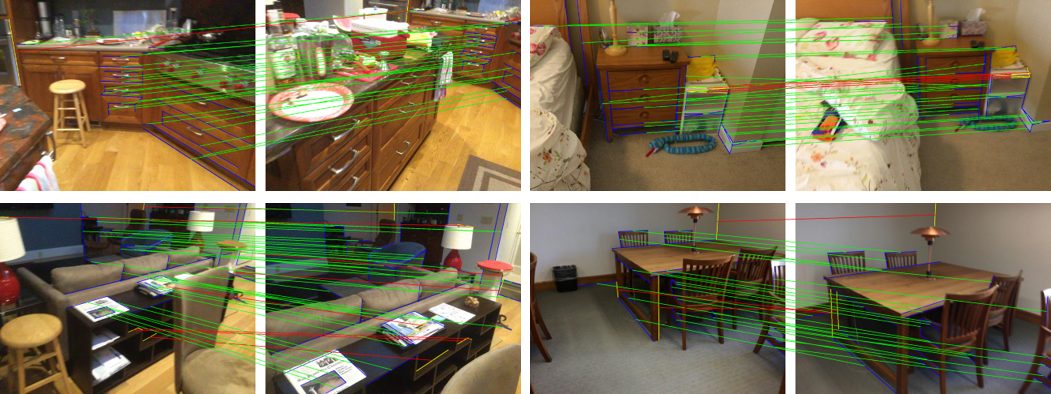}
	\caption{The results generated by our method of four challenging image pairs. }\label{fig-f0}
\end{figure}

However, there are inevitable difficulties in line segments detection. The endpoints of a line segment cannot be accurately located, and the line segments may appear fragmented when the viewpoint changes. These defects may result in variant descriptors for the corresponding line segments from different images. Notably, in low-texture scenes, there is no apparent distinguishable descriptor between the line segments. Therefore, it is challenging to use only the local line segment descriptors to match the correlative line segments from different perspectives. To deal with these challenges, we introduce a novel line segments matching method utilizing graph neural network, which can use the intra-graph and cross-graph convolution to efficiently aggregate global contextual information for robust matching. Fig. \ref{fig-f0} shows the results of our method on four image pairs.

\begin{figure*}
	\centering
	\includegraphics[scale=0.5]{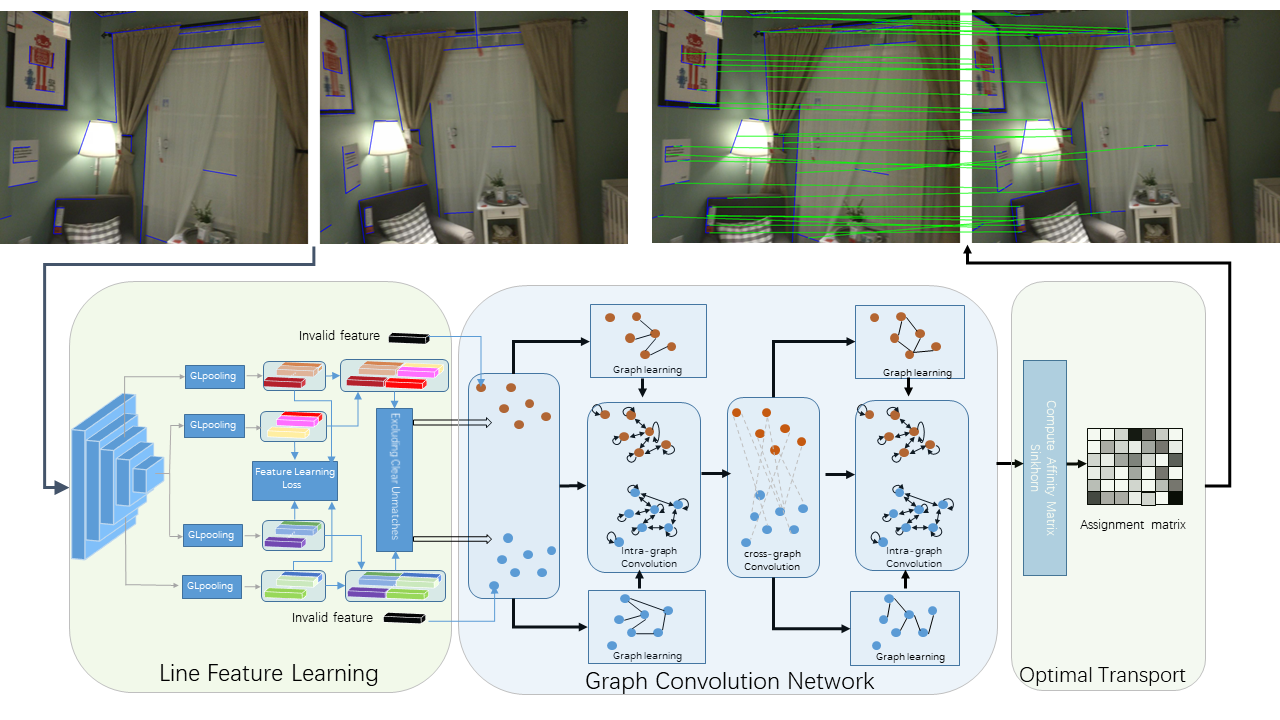}
	\caption{Our network consists of three main modules: Line Feature Learning, Graph Convolution Network and Optimal Transport. Line Feature Learning can learn and extract the line features, then exclude the distinct unmatched lines. Graph Convolution Network includes Intra-graph convolution, Cross-graph convolution and graph architecture learning, which can generate discriminative node embedding features. Optimal Transport computes the affinity matrix between two line sets and the assignment matrix, and uses the assignment matrix to find matches and filter non-matches.} \label{fig-f1}
\end{figure*}

We divided our network into three modules: A line feature learning module is used to learn and compute the local line descriptor. A graph convolution module, including Graph architecture learning, Intra-graph convolution, Cross-graph convolution, is used to learn the global contextual information by context aggregation. And an optimal transport module is used to predict an assignment matrix of two sets. The network is shown in Fig. \ref{fig-f1}. 

The main contributions of this paper can be listed as follows. First, to the best of our knowledge, we are the first to use the Convolutional Neural Networks and Graph Neural Networks to learn the local line descriptor and the matching in a unified end-to-end model to match lines. Second, we design a new line feature extraction algorithm, which extracts the line features from a rectangle centered on the line segment, minimizing the effect of inaccurate locations of the line segment endpoints. Third, we introduce a new graph architecture learning method based on top-k pooling, which is robust to the dataset with unmatched lines. 
\section{Related work}
Line segments matching algorithms have been well studied from different aspects during past decades. The early methods used the geometric constraints to find the correspondence.  
Schmid \cite{schmid2000geometry} utilized epipolar geometry and line segments endpoints to find correlative line segments, and computed the cross-correlation scores of points located on the line segments to match the lines. Hartley \cite{hartley1995linear} used trifocal tensor to match lines between images. However, these methods are sensitive to the changes of line segments endpoints. To acquire better performance, there are some methods grouping adjacent lines to match lines by exploiting the geometry relationship between groups. Shahri \cite{al2014line} used local homography transforms and coplanar relation of adjacent line pairs to detect and verify the candidate correspondences. Li \cite{li2016hierarchical} matched the line segments in groups constructed by adjacent lines and then in individuals by the descriptor of line segments. Wang \cite{wang2009wide} used the spatial proximity to group line segments and then computed the signatures of groups to match lines. Besides, some methods \cite{Lourakis2000Matching, fan2012robust} exploited affine invariant and projective invariant between points and line segments to boost line matching. However, the geometric relationships were hand-crafted and, therefore, sub-optimal. Some methods involve the descriptor of line segment. MSLD \cite{wang2009msld} computed the descriptors by exploiting the mean and standard deviation of the gradients of pixels in a local region of the line segments. SMSLD \cite{verhagen2014scale} was the multi-scale version of MSLD, it can compute scale-invariant line segment descriptor. Bay et al. \cite{bay2005wide} used color histograms to match lines and then eliminated non-matches by a topological filter. Zhang \cite{zhang2013efficient}  used the statistical values of the gradients of the pixels to compute line band descriptor of the line segments and convert the line segments matching problem to graph matching to find matches and eliminate non-matches. Recently, the deep learning methods are used in line segment matching. Vakhitov \cite{vakhitov2019learnable} and DLD \cite{lange2019dld} employed the convolution neural network to adaptively learn the local descriptor of line segments and achieved state-of-the-art results. However, they did not use the contextual information to match lines.

\section{Proposed Approach}
\textbf{Problem description} Consider two images $A$ and $B$, and two sets of line segments ${{\bf{S}}_a}$ and ${{\bf{S}}_b}$ belong to them, respectively. After the images pass through a convolutional neural network ${f_\theta }$, we obtain a stack of convolution maps, where $\theta$ is a trainable parameter. From the convolution maps, we use a line feature extractor $g$ to extract two sets of line segments descriptors ${{\bf{\tilde F}}_a} = \{ {f_1}, \cdots ,{f_{n'}}\}  = g({f_\theta }({\rm A}),{{\bf{S}}_a})$ and ${{\bf{\tilde F}}_b} = \{ {f_1}, \cdots ,{f_{m'}}\}  = g({f_\theta }(B),{{\bf{S}}_b})$, where $n'$ and $m'$ denote the sizes of the two sets, respectively. To reduce the dimension of the line matching problem, some distinct unmatched lines are excluded by a distance criteria between line features. Then we can obtain two sets of filtered line segments descriptors ${{\bf{F}}_a} = \{ {f_1}, \cdots ,{f_n}\}$ and ${{\bf{F}}_b} = \{ {f_1}, \cdots ,{f_m}\}$. In addition, we add an invalid line segment assigned to the unfiltered unmatched line segments. The final two sets of line segments descriptors are ${{\bf{F}}_a} = \{ {f_1}, \cdots ,{f_n},u\}$ and ${{\bf{F}}_b} = \{ {f_1}, \cdots ,{f_m},u\}$, where $u$  is the learnable feature descriptor of the invalid line segments, and the same to learn consistent invalid line feature. So we can construct two graphs ${G_a}({{\bf{F}}_a},{{\bf{A}}_a})$ and ${G_b}({{\bf{F}}_b},{{\bf{A}}_b})$, in which nodes are lines with descriptors ${{\bf{F}}_a}$ and ${{\bf{F}}_b}$ and the graph architecture ${{\bf{A}}_a}$ and ${{\bf{A}}_b}$ are learned adaptively. Our overall goal is to find an optimal assignment matrix ${\bf{P}}$, mapping the nodes from ${G_a}$ to ${G_b}$.
\subsection{Line Feature Learning}
\subsubsection{Backbone Network}
In this paper, we adopt a five-layer VGG-16 \cite{simonyan2014very} network to extract feature descriptors of line segments. In order to obtain low-order and high-order information simultaneously, we extract features from layers 3rd and 5th of the VGG network.
\subsubsection{GLpooing}
This module is used to extract line segments features from the convolutional maps by Gaussian pooling, and we call it GLpooing. To compute the line descriptor, the existing methods  \cite{vakhitov2019learnable}, \cite{zhou2019end} generally extract features from n points located on a line segment, and then a max pooling or mean pooling operator is used to obtain the line segment descriptor. In this way, these methods are not robust to the small changes of line endpoints. Unlike these methods, in this paper, we extract features from Line Support Region by Gaussian pooling. 
\begin{figure}
	\centering
	\includegraphics[scale=0.3]{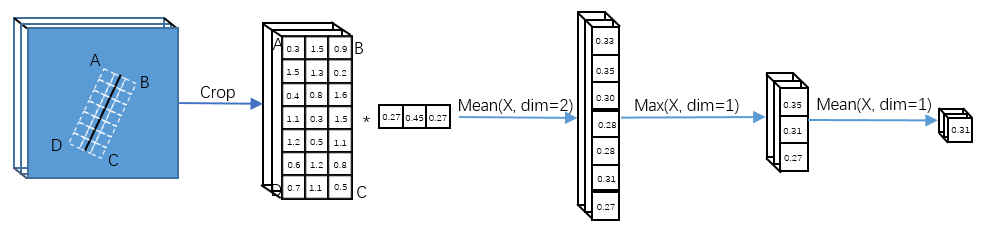}
	\caption{The GLPooling can extract line segments features from the convolutional feature maps. It first crops feature based on the Line Support Region, and then mean or max pooling operator is used to compute line descriptor.} \label{fig-f2}
\end{figure}

The process is shown in Fig. \ref{fig-f2}. Given a line segment, we can draw a rectangle centered on the line segment. Let $m$ be the length of the rectangle, which equals to the length of line segment, and $n$ a hyperparameter reflects the width of the rectangle. We call the rectangle Line Support Region, and use it to crop a feature vector with size $c \times m \times n$ from the convolution maps, where $c$ is the channel dimension of the convolution maps from the backbone network and $m$ is the height, $n$ is the width. Intuitively, the farther away from the line segment, the fewer contributions of the region to the line segment descriptor. To reduce the sensitivity of the line segment descriptor to the small changes of the line segment endpoints, a Gaussian weighting coefficient vector generated by $G(i) = {\textstyle{1 \over {\sqrt {2\pi } \sigma }}}{e^{ - {{d_i^2} \mathord{\left/
				{\vphantom {{d_i^2} {2{\sigma ^2}}}} \right.
				\kern-\nulldelimiterspace} {2{\sigma ^2}}}}}$  is used to entrywise multiply each row of the feature vector, where ${d_i}$ is the distance of the position indexed $i$ to the center. Then, a mean pooling is applied to the feature vector across the width, so the size of the feature vector becomes $c \times m \times 1$. Next, we uniformly divide the feature vector into $w$ sub-vectors across the height and max pool the values in each sub-vector to obtain a feature vector of size $c \times w \times 1$.   Finally, the line segment descriptor is obtained by mean pooling the feature vector along with the height. In this paper, we set $n$ to 7 and $w$ to 5.

\subsubsection{Excluding Distinct Non-matches}
For any line segment, we use the GLPooling to extract the line feature $f$, and regularize it by ${l_{\rm{2}}}$  normalization. To obtain rich semantic descriptors of line segments, we extract line features from the 3rd and 5th layers of VGG for semantically weak and strong features, respectively. Explicitly, the descriptor is ${f} = [f^3||f^5]$, where $f^3$ and $f^5$ are the line features extracted from the 3rd and 5th layer respectively, and $||$ represents concatenation operation. In this way, we can obtain two regularized line descriptors ${f_i} \subset {{\bf{\tilde F}}_a}$ and ${f_j} \subset {{\bf{\tilde F}}_b}$ from images ${A}$ and ${B}$. However, some descriptors in one set can not find the corresponding one from the other set obviously. Let us define a distance $d = \cos {\theta _{ij}} = f_i^T{f_j}$ and a tolerance threshold ${d_s}$, where ${\theta _{ij}}$ is the angle between the descriptors. If the distances of ${f_i}$ in ${{\bf{\tilde F}}_a}$ with all ${f_j}$ in ${{\bf{\tilde F}}_b}$ are all bigger than ${d_s}$, ${f_i}$ is removed, and the line ${\pmb{l}_i}$ in image $A$ is not further considered. The same work is done for lines ${\pmb{l}_j}$ in image $B$.

\subsubsection{Feature Learning Loss}
We adopt Angular Margin Loss as lines feature learning loss, which is a variant of softmax loss  \cite{deng2019arcface}. Due to different layers of VGG have different semantic features, each of the 3rd and 5th layer is associated with a feature learning loss. For a given line segment ${\pmb{l}_i} \subset A$, and its correlative line segment ${\pmb{l}_j} \subset B$, we use the GLPooling to extract the line features from the convolution maps of VGG, and regularize them by ${l_2}$ normalization. The cross product of the regularized line feature $f_i^l$ and $f_j^l$ in layer $l$ is $f_i^{lT}f_j^l{\rm{ = }}\cos \theta _{ij}^l$, where $\theta _{ij}^l$ is the angle between the features. To simultaneously enhance the compactness of matched line segments and the discrepancy of unmatched line segments. An angular margin penalty $\eta^l$ is added between $f_i^l$ and $f_j^l$. So the angular margin loss of between images $A$ and $B$ is defined as
\begin{equation}\label{pb1}
L_{AB}^l =  - {\textstyle{1 \over n}}\sum\limits_{i = 1}^n {\log {\textstyle{{{e^{^{{s^l}\cos (\theta _{ij}^l + {\eta ^l})}}}} \over {{e^{^{{s^l}\cos (\theta _{ij}^l + {\eta ^l})}}} + \sum\limits_{k = 1,k \ne j}^m {{e^{^{{s^l}\cos \theta _{ik}^l}}}} }}}}
\end{equation}
where ${s^l}$ is a hyperparameter. The discriminability of the Eq. (\ref{pb1}) increases as the value of $s^l$ increases. In this paper, we set $s^3$ to 30 and $s^5$ to 5. $\eta^3$ and $\eta^5$ are set to 0.5 and 0.2, respectively. To ensure the balance of two learned line feature sets, the angular margin loss between ${B}$ and ${A}$ is also computed. So the overall loss of the line feature learning is
\begin{equation}\label{pb2}
{L_{feature}} = \sum\nolimits_{l \in \{ 3,5\} } {L_{AB}^l + L_{BA}^l}
\end{equation}

\subsection{Graph Convolution Network}
\subsubsection{Graph Architecture Learning}
Designing a graph structure representing node relationships is important for graph convolution. At present, there are few learning-based methods of line segment matching using graph matching. Delaunary graph and GLMNet \cite{jiang2019glmnet} involved construction graph structure. However, the graph structure designed by them inevitably establish invalid relationships between matched and unmatched line segments in a graph if they are used in line segment matching, because any line segment will establish connections with all adjacency line segments. To overcome this issue, we learn the graph architecture (known as adjacency matrix) ${{\bf{A}}_a}$ and ${{\bf{A}}_b}$ by a top-k pooling to prevent invalid connections from being established.

Given $n+1$ output data features ${{\bf{F}}^l} = \{ f_1^l, \cdots ,f_n^l,{u^l}\}  \in {\mathbb{R}^{(n + 1) \times p}}$ of graph convolution networks from layer $l$. We use a single-layer neural network motivated by \cite{velivckovic2017graph} to compute the relationship score of layer $l+1$. The relation score between node $i$ and node $j$ in a graph is
\begin{equation}\label{pb3}
a_{ij}^l = \sigma ({{\bf{a}}^l}^T[{{\bf{\Omega }}^l}f_i^l|{{\bf{\Omega }}^l}f_j^l])
\end{equation}
where ${{\bf{a}}^l}$ is a learnable parameter, ${{\bf{\Omega }}^l}$ is a learnable weight matrix and $\sigma$ is an activation function, such as RELU. According to the scores, top $\left\lceil {{k^{l + 1}}n} \right\rceil$ adjacent nodes of node i are selected, and the corresponding indices are ${\mathop{\rm idx}\nolimits} _i^{l + 1}$, in which ${k^{l + 1}} \in (0,1]$ is a hyperparameter. To further reduce the invalid pairwise relationships, only when the node $i$ and node $j$ are top $\left\lceil {{k^{l + 1}}n} \right\rceil$ nodes of each other, this relationship would be saved. We define a score matrix ${\bf{A}}_s^{l + 1} \in {\mathbb{R}^{(n + 1) \times (n + 1)}}$ as
\begin{equation}\label{pb5}
{\bf{A}}_s^{l + 1}(i,j) =
\left \{
\begin{array}{rl}
0, \quad & {{\rm{  if   }}j \notin {\mathop{\rm idx}\nolimits} _i^{l + 1}} \\
a_{ij}^{l + 1},\quad & {\rm{  if   }}j \in {\mathop{\rm idx}\nolimits} _i^{l + 1} \\
\end{array}
\right.
\end{equation}
The adjacency matrix is computed by
\begin{equation}\label{pb6}
{{\bf{A}}^{l + 1}} = {\textstyle{1 \over {\left\lceil {{k^{l + 1}}n} \right\rceil }}}\tanh ({\bf{A}}_s^{l + 1}{{{\bf{A}}_s^{l + 1}}^T})
\end{equation} 
where  $\left\lceil {{k^{l + 1}}n} \right\rceil$ is used to approximatively normalize each row of adjacency matrix. Intuitively, there should be no relationship between every detected line and invalid line, so each value of the last row and the last column of ${{\rm{A}}^{l + 1}}$ is set to zero.

Generally, compared with the higher layers, it is more difficult for the lower layers to distinguish the matched and unmatched nodes, so the lower layers should keep more neighbors. In this paper, the ${k^l}$ is computed by ${k^l} = \max ({\textstyle{{{\rm{0}}{\rm{.4}}} \over {{{\rm{2}}^l}}}},0.1)$.
\subsubsection{Intra-graph Convolutional module}
Intra-graph convolution aggregates features from the adjacent unambiguous nodes and the node itself to generate discriminative global embedding node features and discard ambiguities. A large number of graph convolution operators can be chosen from recent works  \cite{velivckovic2017graph,kipf2016semi,gilmer2017neural,schlichtkrull2018modeling,xu2018powerful,wang2019learning}.  In this paper, we use the graph convolution operator used in \cite{wang2019learning} for its simple implementation
\begin{equation}\label{pb7}
{{\bf{F}}^{l + 1}} = \sigma ({{\bf{A}}^l}{{\bf{F}}^l}{\bf{\Theta }}_1^l) + \sigma ({{\bf{F}}^l}{\bf{\Theta }}_2^l)
\end{equation} 
where ${\bf{\Theta }}_1^l$ and ${\bf{\Theta }}_2^l$ are learnable parameters, $\sigma$ is the activation function RELU.
\subsubsection{Cross-graph Convolution Module}
Cross-graph convolution further improves the similarity of global embedding node features of the matched node by aggregating features from nodes with similar features in the other graph. Given the input features ${{\bf{F}}_a}$ and ${{\bf{F}}_b}$ of both graphs, we can obtain the output node embedding features ${\bf{F}}_a^l$ and ${\bf{F}}_b^l$ of layer $l$ from the shallow layer, then we can utilize this node embedding features to compute the soft assignment matrix ${\rm{P}}_{ab}^l \in {^{(n + 1) \times (m + 1)}}$  by optimal transport layer, in which the $i$-th row vector of ${\rm{P}}_{ab}^l$ can be considered as correspondence scores between node $i$ in graph $G_a$ and each node in graph $G_b$. The Cross-graph convolution can be conducted as
\begin{equation}\label{pb11}
{\bf{F}}_a^{l + 1}(1:n,:) = [{\bf{P}}_{ab}^l(1:n,:){\bf{F}}_b^l||{\bf{F}}_a^l(1:n,:)]{{\bf{W}}^l}
\end{equation} 
\begin{equation}\label{pb12}
{\bf{F}}_b^{l + 1}(1:m,:) = [{\bf{P}}_{ab}^l{(1:m,:)^T}{\bf{F}}_a^l||{\bf{F}}_b^l(1:m,:)]{{\bf{W}}^l}
\end{equation} 
\begin{equation}\label{pb13}
{\bf{F}}_a^{l + 1}(n + 1,:) = {\bf{F}}_a^l(n + 1,:)
\end{equation} 
\begin{equation}\label{pb14}
{\bf{F}}_b^{l + 1}(m + 1,:) = {\bf{F}}_b^l(m + 1,:)
\end{equation} 
Where ${{\bf{W}}^l}$ is the trainable weight matrix. Invalid nodes do not aggregate embedding features from another graph because all unmatched line segments from the other graph are assigned to it.
\subsection{Optimal Transport Layer}
After obtaining the embedding features of nodes via the graph convolution layer, the matching of two graphs can be interpreted as the node-to-node affinity metric. So the optimal transport layer can simultaneously perform the matching and filtering process. Given the output embedding features ${{\bf{F}}_a}$ and ${{\bf{F}}_b}$ of two graphs, the affinity matrix $\bf{M}$ can be calculated by
\begin{equation}\label{pb15}
{\bf{M}} = \exp ({\textstyle{{{{\bf{F}}_a}{\bf{CF}}_b^T} \over \delta }})
\end{equation} 
Where $\bf{C}$ is a learnable weight matrix, $\delta$ is a hyper parameter. The discrimination of $\bf{M}$ increases with the decrease of $\delta$. 

The line matching problem has two constraints: First, the matched lines should have exactly single correspondence in the other image. Second, the invalid line will be assigned to all unmatched lines in the other image. The soft assignment matrix ${\bf{P}} \in {[0,1]^{(n + 1) \times (m + 1)}}$, representing the correspondences of two line segments sets, will be computed by solving the optimization problem  $\max \sum\limits_{i = 1}^{n + 1} {\sum\limits_{j = 1}^{m + 1} {{{\bf{M}}_{ij}}{{\bf{P}}_{ij}}} } $ under these two constraints.

\begin{equation}\label{pb16}
{\bf{P}}{{\bf{1}}^{m + 1}} = {\bf{a}}
\end{equation}
\begin{equation}\label{pb17}
{{\bf{P}}^T}{{\bf{1}}^{n + 1}} = {\bf{b}}
\end{equation}
Where ${{\bf{1}}^{m + 1}}$ and ${{\bf{1}}^{n + 1}}$ are vectors with dimensions ${m+1}$ and ${n+1}$, respectively, and their elements are $1$, the vector ${\bf{a}}$ and ${\bf{b}}$ are $[{{\bf{1}}^n},m]$ and $[{{\bf{1}}^m},n]$, respectively.

This constrained optimization problem can be considered as an optimal transport problem \cite{peyre2019computational}. The soft assignment matrix ${\bf{P}}$ can be solved in differentiable, parallel ways by the Sinkhorn algorithm \cite{sinkhorn1964relationship} on GPU.

\begin{figure*}
	\centering
	\subfigure[The results of effect of rotation, blurring and scale on presicion]{
		\begin{minipage}{0.33\linewidth}
			\centering
			\includegraphics[width=1.1\textwidth]{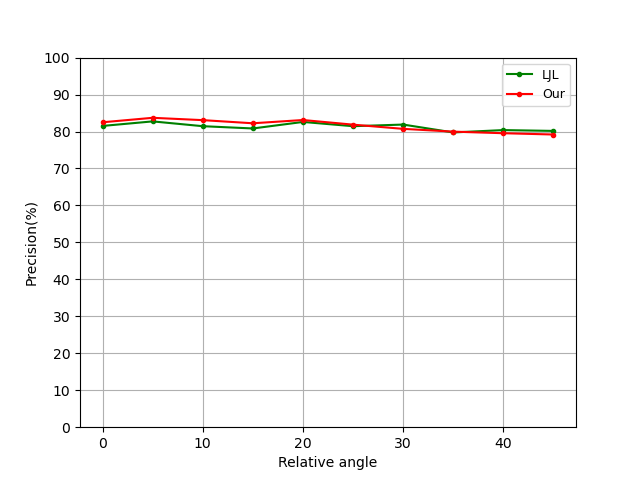}
		\end{minipage}%
		\begin{minipage}{0.33\linewidth}
			\centering
			\includegraphics[width=1.1\textwidth]{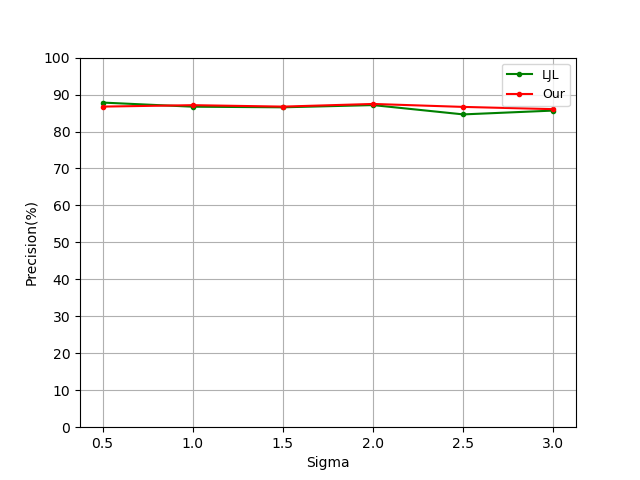}
		\end{minipage}%
		
		\begin{minipage}{0.33\linewidth}
			\centering
			\includegraphics[width=1.1\textwidth]{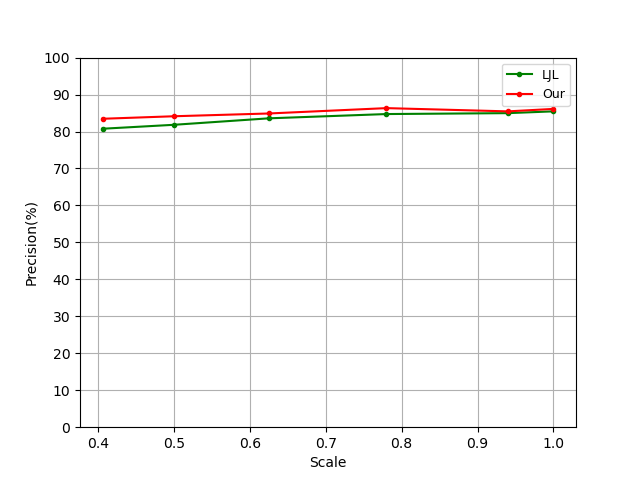}
		\end{minipage}
		
	} \\ %
	\subfigure[The results of effect of rotation, blurring and scale on recall]{
		\begin{minipage}{0.33\linewidth}
			\centering
			\includegraphics[width=1.1\textwidth]{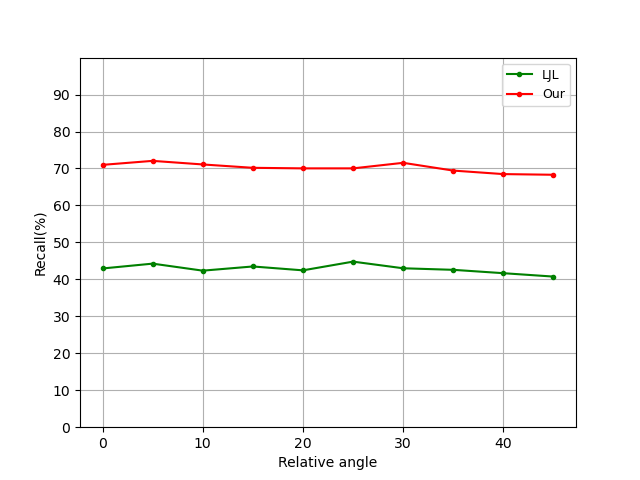}
		\end{minipage}%
		\begin{minipage}{0.33\linewidth}
			\centering
			\includegraphics[width=1.1\textwidth]{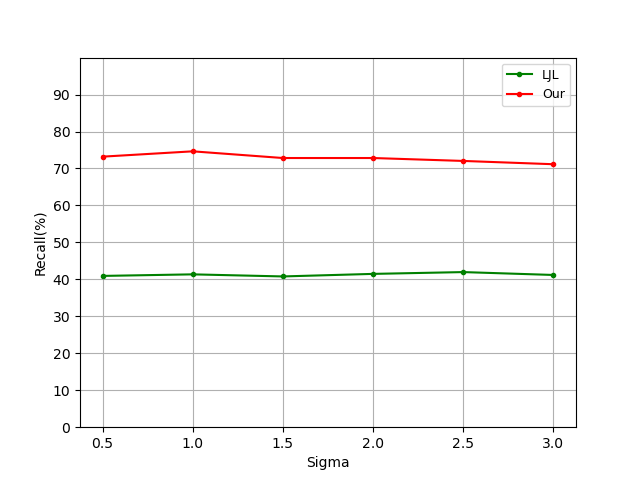}
		\end{minipage}
		\begin{minipage}{0.33\linewidth}
			\centering
			\includegraphics[width=1.1\textwidth]{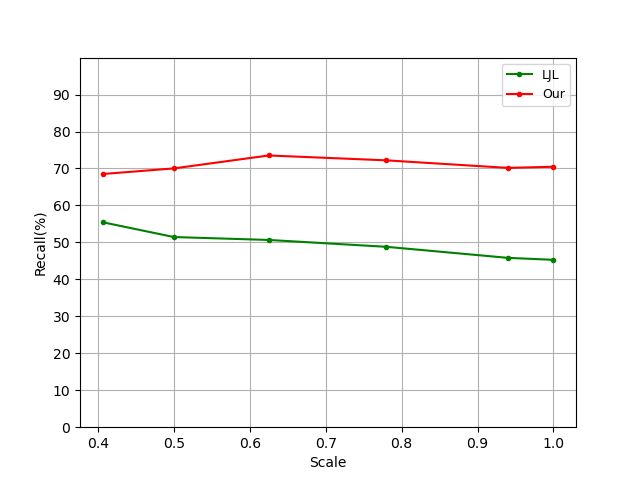}
		\end{minipage}%
	}
	
	\centering
	\caption{ The comparison of the presicion and recall between our method and LJL. We rotate two images in opposite directions at the same time. The presicion and recall with respect to the Relative angles between two images are shown in the left. The presicion and recall with respect to the standard deviation of gaussian blur are shown in the middle, where Sigma is from 0.5 to 3. The presicion and  recall with respect to the Scale of images shown in the right, where scale is changed from 0.4 to 1, and the image size is 1024*764 when scale is 1. 
	}\label{fig-f3}
\end{figure*}
We use the negative log-likelihood loss as the matching prediction loss. The ground truth is $\mathcal{{\cal G}{\cal T}}$, which includes all matched lines index tuples $(i ,j)$, $\mathcal{{\cal A}}$, $\mathcal{{\cal B}}$ are index sets of unmatched line segments of images $A$ and $B$, respectively. The matching prediction loss is
\begin{align}\label{pb18}
{L_{graph}} =  &- \sum\limits_{(i,j) \in \mathcal{{\cal G}{\cal T}}} {\log {{\bf{P}}_{i,j}}} \notag 
- \sum\limits_{i \in \mathcal{{\cal A}}} {\log {{\bf{P}}_{i,m + 1}}}  \\
&- \sum\limits_{j \in \mathcal{{\cal B}}} {\log {{\bf{P}}_{n + 1,j}}}
\end{align}
So, the overall loss of the network is
\begin{equation}\label{pb19}
L = \lambda {L_{feature}} + (1 - \lambda ){L_{graph}}
\end{equation}
Where $\lambda$ balances two terms and is set to 0.5 in our experiments.
\section{Experiments and Results}
\subsection{Training details}
To train proposed network and evaluate its performance, vast amounts of data with ground truth labels are needed. We collect large quantities of images from ScanNet \cite{dai2017scannet}, which is a large-scale indoor dataset with color images, depth images and poses from different indoor scenes. Manually labeling ground truth labels for a pair of images is inefficient and tedious. To solve this problem, in this paper, we use a method to find line segments matches from a pair of images. In short, we use LCNN \cite{zhou2019end} to detect line segments from both images. Then, for any line segment ${\pmb{l}_a}$ in one image $A$, we back project it to the depth image to locate a 3D line segment ${\pmb{L}_a}$. And project ${\pmb{L}_a}$ to the other image $B$ as line segment ${\pmb{l}'_a}$. If ${\pmb{l}'_a}$ is near to a line segment ${\pmb{l}_b}$ on the image $B$, the angle between ${\pmb{l}'_a}$ and ${\pmb{l}_b}$ is less than a threshold, and the overlap of them is greater than another threshold, we label ${\pmb{l}_a}$ and ${\pmb{l}_b}$ is a pair of matches. However, since the value of the depth image is not very accurate, we can only guarantee the ratio of matches to non-matches is from 0.5 to 1.5. We also discarding some pairs of images with too small matches or too large overlap. Finally, the training set includes top 590 scenes of Scannet and 25000 pairs, the left is test set, which includes 6000 pairs. In this paper, we use the Adam optimizer \cite{kingma2014adam} to train network with initialing learning rate 1e-3 and decay the learning rate by 10 on each epoch. The batch size is set to 4, we stop the training at 10 epochs as the loss converges. The training spends 4 hours on a single NVIDIA GTX 2080Ti GPU. During the training, images are randomly resized and rotated.

\subsection{Performance evaluation}
\textbf{Quantitative comparison} To evaluate the performance of our network, we compare the performance of our method with several line matching methods (MSLD \cite{wang2009msld}, LPI \cite{fan2012robust}, LJL \cite{li2016hierarchical}) with outstanding performances, which also have no special requirements for line extraction algorithms. We quantitatively evaluate the performance of different methods using Presicion (P), Recall (R) and ${{\rm{F}}_{{\rm{measure}}}} = {\textstyle{{2{\rm{RP}}} \over {{\rm{R}} + {\rm{P}}}}}$.  Due to the implementation of LJL provided by its authors cannot find matches in some images in the test set, for a fair comparison, we compare the performance on the images which can be found matches by LJL, and the results are shown in Table \ref{tab-b1}. From the table, we can conclude that our method achieves significantly better performance compared with these line matching methods. The presicion of our method is slightly higher than LJL. Better, our method achieves 70.47\% recall, which is 25\% more than that of LJL. It represents our method can produce more correct matches and is more robust to various indoor scenes.

\begin{figure*}
	\centering
	\begin{minipage}{1.0\linewidth}
		\centering
		\includegraphics[width=1.0\textwidth]{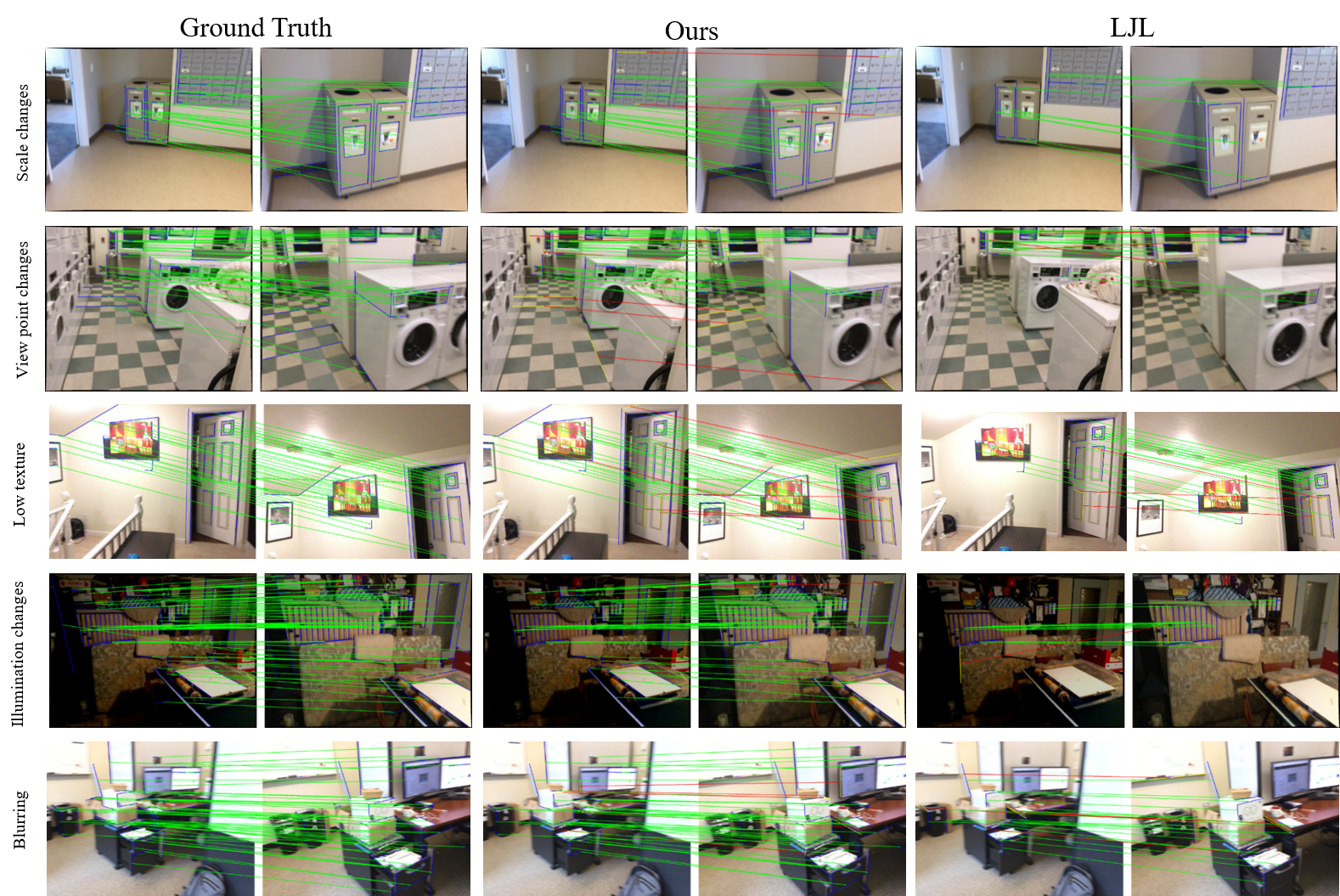}
		\caption{Qualitative evaluation of our line segments matching method. We compare our method with LJL. Left: ground truth. Middle: our prediction. Right: the prediction of LJL. Our method finds more matches and works well in the scenes of view point changes, scale changes, low texture, illumination changes and blurring.
		}\label{fig-f4}
	\end{minipage}%
	
\end{figure*}

\begin{table}
	\caption{The presicion, recall and ${{\rm{F}}_{{\rm{measure}}}}$ of four methods.}\label{tab-b1}
	\begin{tabular}{|c|c|c|c|}
		\hline
		Methods & Presicion (\%) & Recall (\%) & ${{\rm{F}}_{{\rm{measure}}}}$ (\%) \\
		\hline
		LJL & 85.46 & 45.29 & 59.20 \\
		\hline
		LPI & 69.28 & 20.12 & 31.18 \\
		\hline
		MSLD & 69.29 & 29.22 & 41.11 \\
		\hline
		Our Method & 86.12 & 70.47 & 77.51\\
		\hline
	\end{tabular}
\end{table}

We also select and process some images from the test set to compare the robustness of our proposed algorithm with LJL in terms of image transformation, including rotation changing, scale changing, and image blurring. The reason we don't compare with LPI and MSLD is their presicion and recall are much lower than LJL and our method. The results are shown in Fig.\ref{fig-f3}. We rotate two images in opposite directions at the same time. The presicion and recall with respect to the relative angles between two images are shown in the left. The presicion and recall with respect to the standard deviation of Gaussian blur are shown in the middle, where Sigma is from 0.5 to 3. The presicion and recall with respect to the scale of images shown in the right, where scale is changed from 0.4 to 1, and the image size is 1024*764 when scale is 1.  From the perspective of presicion, rotation and blurring have little effect on LJL and our method. Both methods have good results for scale changes. But in detail, the presicion of our method is slightly better than that of LJL. However, the recall of our results are much better than that of LJL no matter in rotation changing, scale changing or image blurring. In Fig. \ref{fig-f3}(c), we notice that the recall of LJL at the small scale is higher than that at the large scale. The reason is that the LJL generates more junctions to match lines through groups in small scale by using the same threshold.

\textbf{Ablation studies} In order to evaluate the performance of three main components: Feature learning loss, Graph learning by top-k pooling and GLpooling in our network, we design ablation experiments with different components combination. The scale is set to 0.5 in the experiments. If the Graph learning by top-k pooling is absent, the method proposed in \cite{jiang2019glmnet} is used. If the GLPooling is absent, the points located on line segments is used to compute the line descriptors. The results are reported in Table \ref{tab-b2}. We can notice that the proposed graph architecture learning method improves the performance of the results, which indicates that it can prevent establishing some invalid relationships and is more beneficial for the dataset with unmatched lines. GLpooling is more robust to the small changes of line segment endpoints and is useful for computing invariant line descriptors. And the added feature learning loss improves the performance of line segments matching by simultaneously enhancing the compactness of matched line segments and the discrepancy of unmatched line segments.

\begin{table}[!t]
	\centering 
	\caption{Result of ablation studies on entire test dataset.}\label{tab-b2}
	\setlength{\tabcolsep}{0.51mm}{
	\begin{tabular}{|c|c|c|c|c|}
		\hline
		\tabincell{c}{Feature Learning\\ Loss} & \tabincell{c}{Graph learning \\by top-k pooling} & GLpooling & P& R\\
		\hline
		$\surd$ & $\surd$ & $\surd$ & 81.27 & 69.46\\
		\hline
		$\surd$ & $\surd$ & $\times$ & 80.78 & 68.06\\
		\hline
		$\surd$ & $\times$ & $\times$ & 79.27 & 66.83\\
		\hline
		$\times$ & $\times$ & $\times$ & 76.83 & 57.73 \\
		\hline
	\end{tabular}}
\end{table}

\textbf{Qualitative analysis} Some visualized results of our method and LJL are shown in Fig. \ref{fig-f4}. In the figure, we mark matched lines in blue and unmatched lines in yellow. Green lines identify correct pairs and red lines are incorrect pairs. It can be seen that our method generates more correct matches and successfully copes with view point changes, scale changes, low texture, illumination changes and blurring.

\section{Conclusion}
We introduce a novel line segments matching model by exploiting the convolutional neural networks and graph neural networks, our method can learn the local line descriptor and the matching in a unified end-to-end model. In this paper, a new line feature extraction algorithm is introduced to extract line features with considering the inaccurate locations of the line segment endpoints, a new learnable graph architecture based on top-k pooling is used to prevent to establish invalid connections, and adding the invalid line segments feature is used to deal with the unmatched lines. Experiments show that the method can achieve satisfactory and robust performance compared with the state of the art methods.

\bibliographystyle{cas-model2-names}

\bibliography{cas-refs}


\end{document}